\documentclass[letterpaper, 10 pt, conference]{ieeeconf}  

\IEEEoverridecommandlockouts                              

\usepackage[top=0.75in, left=0.75in, right=0.75in, bottom=0.76in]{geometry}

\usepackage{graphics} 
\usepackage{epsfig} 
\usepackage{mathptmx} 
\usepackage{times} 
\usepackage{amsmath} 
\usepackage{amssymb}  
\usepackage{graphicx}
\usepackage{cite}
\usepackage{xcolor}
\title{\LARGE \bf
\textcolor{black}{A Quadrotor with an Origami-Inspired  Protective Mechanism}
}

\author{Jing Shu and Pakpong Chirarattananon%
\thanks{This work was partially supported by the Research Grants Council of the Hong Kong Special Administrative Region of China (grant number CityU-21211315).}
\thanks{J. Shu is with the Department of Mechanical Engineering and P. Chirarattananon is with the Department of Biomedical Engineering, City University of Hong Kong, Hong Kong SAR, China (email: pakpong.c@cityu.edu.hk).}%
}

\begin{document}

\maketitle
\thispagestyle{empty}
\pagestyle{empty}

\begin{abstract}
\textcolor{black}{Despite advances in localization and navigation, aerial robots inevitably remain susceptible to accidents and collisions.} In this work, we propose a passive foldable airframe as a protective mechanism for a small aerial robot. A foldable quadrotor is designed and fabricated using the origami-inspired manufacturing paradigm. Upon an accidental mid-flight collision, the deformable airframe is mechanically activated. The rigid frame reconfigures its structure to protect the central part of the robot that houses sensitive components from a crash to the ground. The proposed robot is fabricated, modeled, and characterized. The 51-gram vehicle demonstrates the desired folding sequence in less than 0.15 s when colliding with a wall when flying.
\end{abstract}

\section{Introduction}
Recent rapid developments of aerial robots have shown promise. Following the advances in flight dynamics and control \cite{mulgaonkar2018robust}, planning and localization \cite{qin2017vins,chirarattananon2018direct}, etc., there emerge numerous applications of these small flying robots that involve interactions of robots with objects or environments. These include, for instance, transportation of a suspended payload \cite{lee2018geometric}, climbing on a vertical surface \cite{pope2017multimodal}, perching on an overhang \cite{graule2016perching, hsiao2018ceiling}, aerial manipulation \cite{kim2018origami}. As the complexity of the tasks grows, it inevitably escalates the chance of failures. Despite attempts to circumvent an accident, it is still likely unforeseen circumstances would lead to an undesired collision that destabilizes the flight. The impact from a subsequent fall could lead to a destructive damage on the robot.

Researchers have proposed various strategies to deal with collisions. Thus far, the most common direction is to reduce the detrimental impact from collisions so that the robots retain the attitude stability. One solution is to incorporate a protective frame \textcolor{black}{that elastically absorb impact energy \cite{klaptocz2013euler}} or is dynamically decoupled from the robot's body \cite{briod2014collision, sareh2018rotorigami}. With an appropriate damping mechanism, this allows the robot to continue flying as the influence on the attitude dynamics is drastically reduced. Another approach is to develop vehicles that are mechanically and dynamically robust. The flapping-wing robot in \cite{tu2019acting}, for example, is capable of navigating tight space as it is resilient against collisions owing to the intrinsic compliance of its aerodynamic surfaces. Alternatively, in \cite{mintchev2017insect},\textcolor{black}{\cite{mintchev2018bioinspired}}, the authors opt to embrace the collisions and ensure the robot suffers no damage from the succeeding fall. The insect-inspired collision tolerance is accomplished with a deformable airframe that is held rigid by the magnetic joints \cite{mintchev2017insect} or prestretched elastomeric membranes \cite{mintchev2018bioinspired}. The energy absorbing property of the material shelters the central case from a violent impact.

\begin{figure}[t]
\centering
\includegraphics[width=8.4cm]{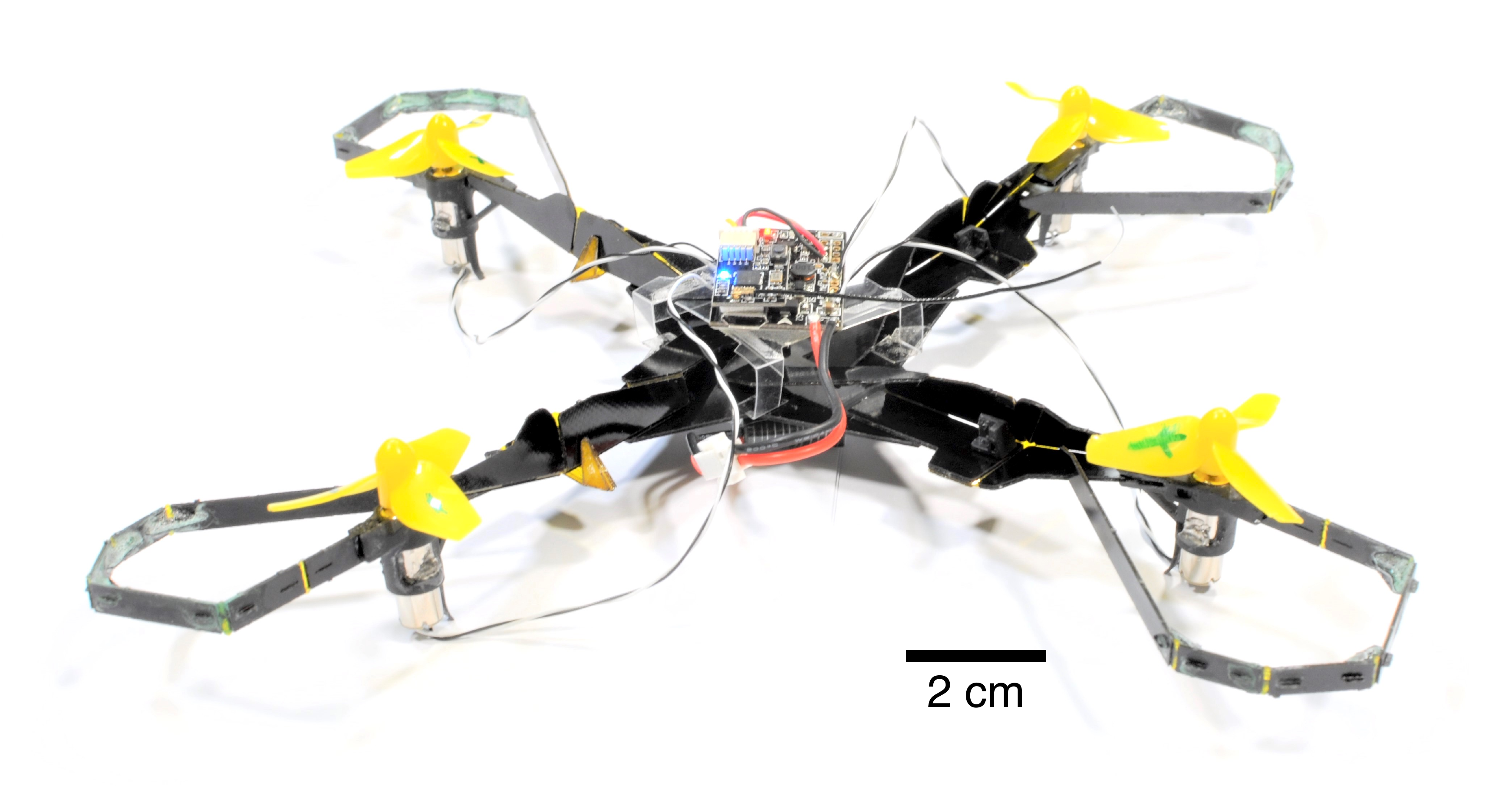}
\caption{Photo of a 51.2-gram foldable quadrotor fabricated by the origami-inspired method \cite{rus2018design,zuliani2018minimally}. The fold is mechanically activated upon a mid-flight impact to a vertical surface.}
\label{fig:robot_prototype}\vspace{-2mm}
\end{figure}

This paper addresses the issue of destructive falls from collisions of aerial robots by taking an inspiration from the conglobating behavior of pill bugs \cite{smigel2008conglobation},  Armadillidiidae and pill millipedes exhibit a defensive mechanism when triggered by stimuli by rolling into a ball. In conglobation, legs and sensitive ventral surfaces are wrapped and protected by the segmented dorsal exoskeleton. Herein, the observed defensive strategy is translated into a deformable multicopter in Fig. \ref{fig:robot_prototype} that folds the rigid airframe to safeguard central components when triggered by a collision. As a result, the delicate parts are shielded from the impact in the subsequent drop to the ground.

Unlike the approach in \cite{mintchev2017insect},\textcolor{black}{\cite{mintchev2018bioinspired}}, which employs a synergic implementation of a dual-stiffness behavior and energy absorbing structures, we implement an origami-inspired method\cite{rus2018design,zuliani2018minimally} to create a transformable airframe for collision tolerance.  \textcolor{black}{The morphological adaptation has demonstrated various functions in aerial robots, including for protection and safety, \cite{kornatowski2017origami,sareh2018rotorigami}, improved agility \cite{stowers2015folding,riviere2018agile}, rapid deployment \cite{henderson2017towards}, and aerial grasping \cite{ kim2018origami}.} Upon a collision, our robot deforms by folding according to the predetermined features without the need of extra sensors or actuators. To achieve the desired mechanical intelligence, the foldable arms are built to be nominally rigid in the flight configuration. In addition, the interlocking mechanism for transitioning between the flight and folded states is purely mechanical. Both aspects potentially reduce the weight of the robot.

In the next section, we elaborate on the design principles and describe how the structural rigidity is attained from flat parts. Important flight components and the fabrication method are given. The section describes the fold kinematics and the mechanism for transitioning between the flying state and folded state. Section \ref{sec:benchtop_experiments} sets out the experimental characterization of the proposed robot and compares the results to the model predictions. Flight and mid-air collision are presented in Section \ref{sec:mid-flight_collision}, followed by conclusion and discussion.

\section{Foldable Quadrotor Design and Model}\label{sec:design_and_model}
\subsection{Overview}

\begin{figure}[t]
\centering
\includegraphics[width=8cm]{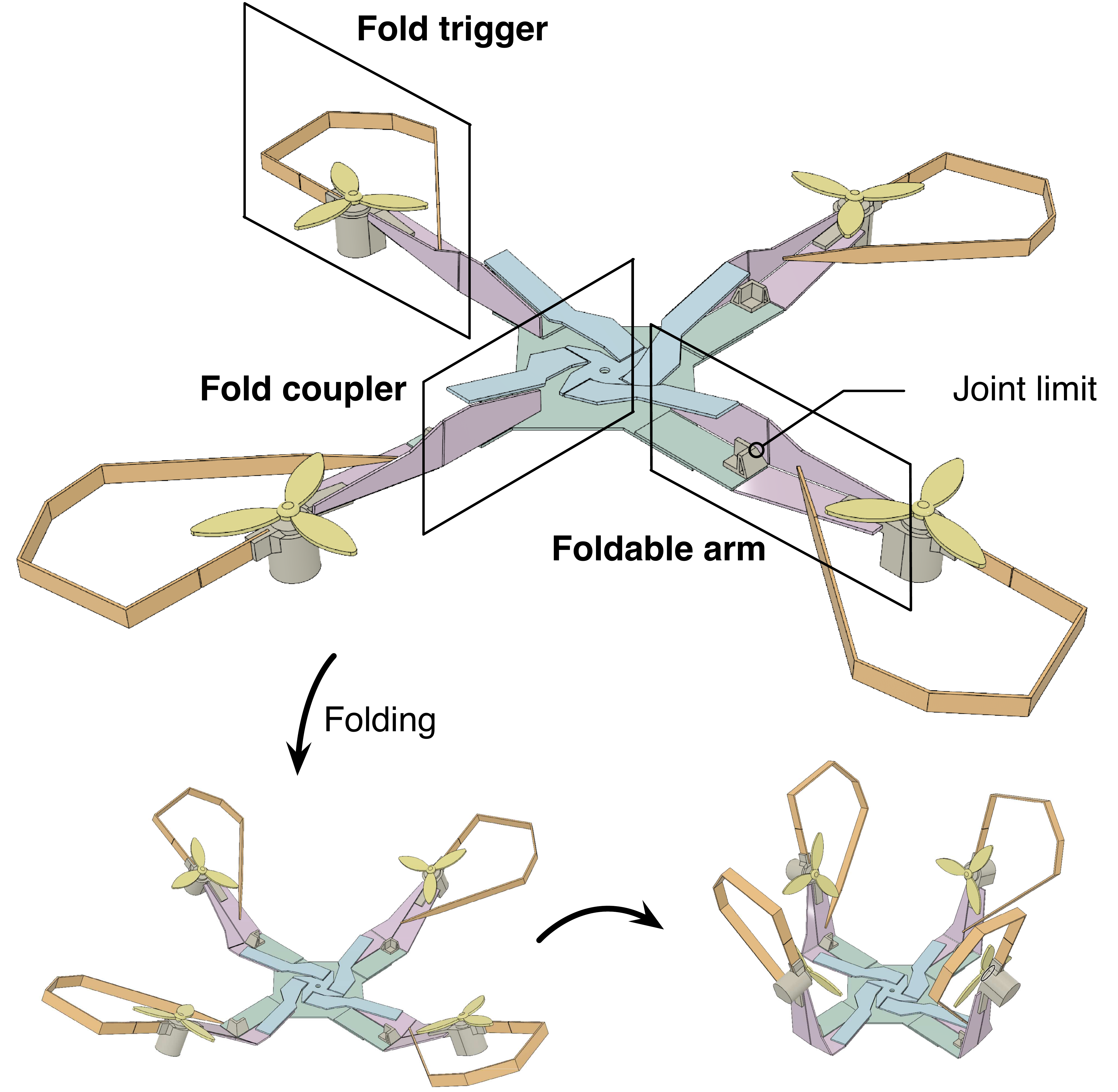}
\caption{\textcolor{black}{An illustration of the three major components (foldable arm, fold coupler, and fold trigger), joint limits, and the folding sequence.}}
\label{fig:panel_folding_cad}
\end{figure}
\begin{figure}[t]
\centering
\includegraphics[width=8cm]{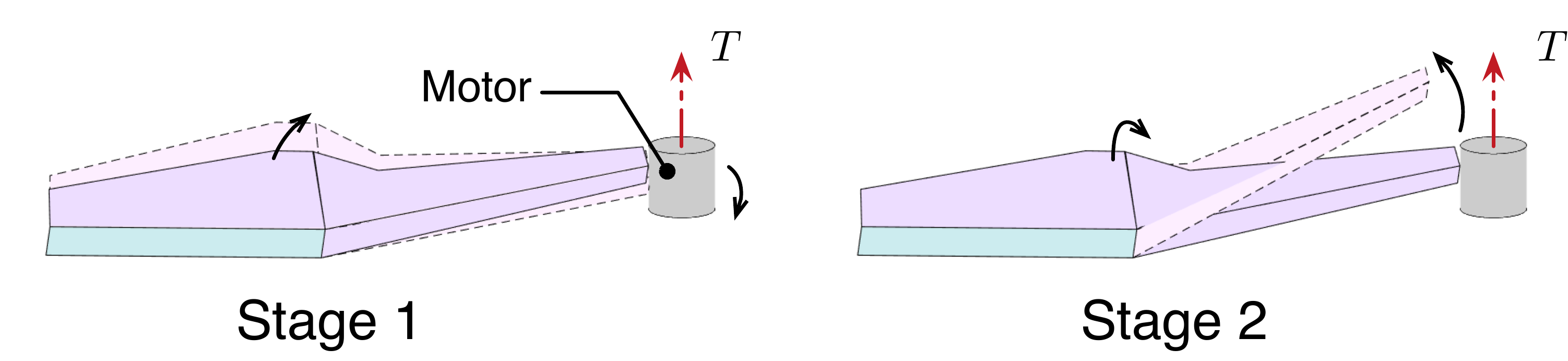}
\caption{\textcolor{black}{A diagram illustrating two folding stages of the airframe upon an in-flight collision. From the flight configuration, the motor is initially displaced downwards and later shifted up according to the predetermined kinematics.}}
\label{fig:panel_folding_stages}
\end{figure}
In the mechanical design process, we aim to create an aerial robot that is mechanically resilience to withstand an impact. This must be achieved while preserving the structural rigidity and keeping the mass minimal as required for flight.

In the conglobation of pill bugs \cite{smigel2008conglobation}, it can be seen that several rigid  shells in multiple segments shift from an approximately planar alignment to form a circular arc. While each segment keeps its intrinsic stiffness, the deformation is globally achieved by small adjustments of the relative orientations. This is akin to how a large structural shift can be obtained from folding flat rigid structures. The observation provides a motivation to employ the origami-inspired design to realize the analogous robotic conglobation with aerial robots. 

In contrast to the designs in \cite{sareh2018rotorigami, zhao2018deformable}, our foldable design is relatively simple and minimal. Each feature is clearly customized for the desired property. First, we take the minimum amount of folding joints to ensure the airframe is rigid in the desired directions when flying, yet foldable when triggered. Second, the design allows the folds to be mechanically activated upon a collision. In addition, all the folds are coupled as a one degree of freedom system. All arms fold in synchronization for the protection of the central body.

Fig \ref{fig:panel_folding_cad} schematically demonstrates the folding mechanism of the proposed robot. The foldable airframe of the prototype consists of the ground tile, foldable arms, fold coupler, and fold triggers. In the flying state, the propeller's thrust is nominally vertical. Each propeller generates an upward force, of which the resultant torque is countered and balanced by the joint limits, keeping the arms in the flight configuration. Upon a collision, the impact rotates the fold trigger to push part of the arms. \textcolor{black}{If the force is sufficiently large, it overcomes the torque contributed by the propellers' thrust about the fold axis, activating the arm folding motion as shown in Fig. \ref{fig:panel_folding_cad}}.

\subsection{Components and Fabrication}
The origami-inspired quadrotor consists of standard electronic parts and the foldable airframe. We employ commercial off-the-shelf parts: 7$\times$20-mm brushed motors with a rated no-load speed of 54,000 RPM at 3.7V, 40-mm 3-blade propellers, and a single-cell 400-mAh Li-ion battery. Together, four propellers can generate over 0.65 N. A F3-EVO brushed flight controller is used for stabilization and receiving user's commands through a mini radio receiver. 

The origami-inspired airframe was manufactured from the planar fabrication paradigm \cite{rus2018design,zuliani2018minimally}. Sheets of materials were cut and patterned using CO$_2$ laser (Epilog Mini 24). The laser-cut layer of polyimide film (25-$\mu$m Kapton, Dupont) was sandwiched between two structural layers (300-$\mu$m fiberglass). Parts and features were pin aligned and 170-$\mu$m double-sided pressure sensitive adhesives (300LSE, 3M) were used to compose the laminates as shown in Fig. \ref{fig:panel_folding_arm}A. The laminates were released from the frame by the CO$_2$ laser in the final steps. The use of middle flexible layer enables us to create flexural (revolute) joints, while the structural rigidity is provided by the fiberglass. Folding the planar laminates gives rise to 3D assemblies. 

The airframe was constructed from three laminates \textcolor{black}{as illustrated in Fig. \ref{fig:panel_folding_arm}A. The first constitutes the robot's base, acting as a mechanical ground. The second laminate makes up the foldable arms and the coupler. The third is the fold triggers, reinforced by 200-$\mu$m transparency film for increased joint stiffness}. The foldable airframe is then fitted with joint limits, motors, propellers, and flight electronics aided by 3D printed components (Black Resin, Form 2, Formlabs). In total, the robot's mass is 51.2 g. The mass of the foldable airframe, including the mechanical ground, arms, and fold triggers, is 14.9 g.

\subsection{Foldable Arms}

\begin{figure}[t]
\centering
\includegraphics[width=8cm]{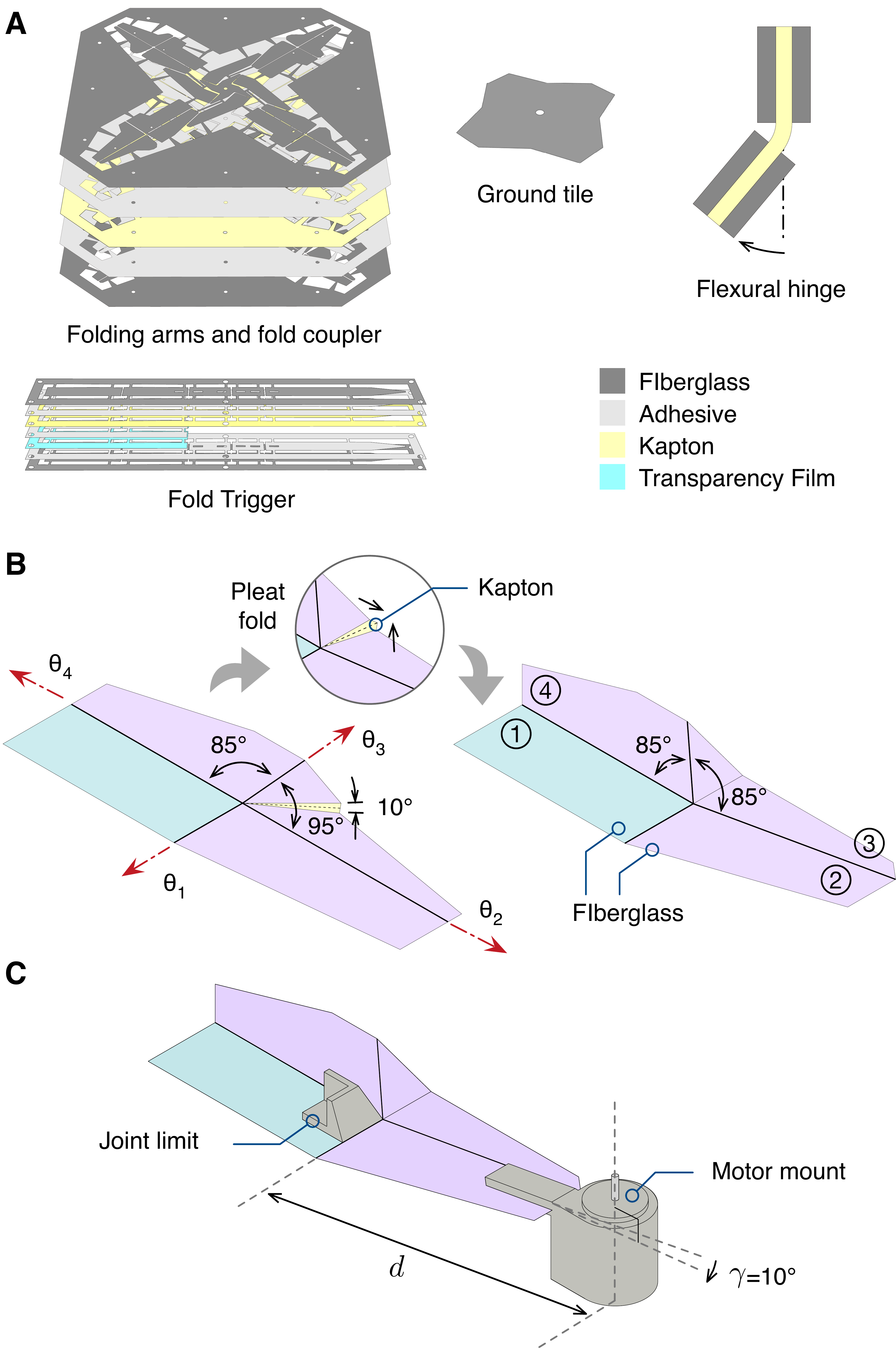}
\caption{(A) \textcolor{black}{Laminates of fiberglass, Kapton sheets, adhesive, and transparency film for the planar fabrication of the foldable airframe, the ground tile, fold triggers and a resultant flexural joint (not to scale).} (B) The conceptual fabrication and the joint kinematics of the foldable arm. (C) The added components on the airframe and the associated variables.}
\label{fig:panel_folding_arm}\vspace{-2mm}
\end{figure}

The airframe plays the most vital role in the folding mechanism. As in other flying robots, the flight condition requires the structure to be rigid in multiple directions. To satisfy this stringent restriction with the limitation of thin planar structures (which are relatively compliant in one direction as the stiffness is proportional to the cube of the thickness), we incorporate folds into the design. 
\textcolor{black}{The fold kinematics are designed such that, upon a collision, the airframe reconfiguration can be described in two stages as depicted in Fig. \ref{fig:panel_folding_stages}. At the beginning, the collision force is redirected to  shift  the  motors  downwards  or  in  the  direction  opposite to the propelling thrust. The motion is then reversed and the motors  and  airframe fold  up, forming a protective structure. During the first stage, the impact energy is absorbed as the work done against the propellers and temporarily stored in the mechanical structure rather than in elastic materials as found in \cite{graule2016perching,mintchev2017insect,mintchev2018bioinspired}. To this end, our foldable design relies entirely on rigid structures and flexural hinges, eliminating the use of viscoelastic components for energy storage. The  approach simplifies the fabrication and modeling effort.}

As seen in Fig. \ref{fig:panel_folding_cad}, in the flying configuration, each foldable arm consists of pairs rigid tiles oriented approximately in perpendicular. In combination, this prevents each arm from bending up owing to the propeller's thrust, or from deforming about the vertical axis due to the propeller's yaw torque.

\begin{figure}[t]
\centering
\includegraphics[width=8cm]{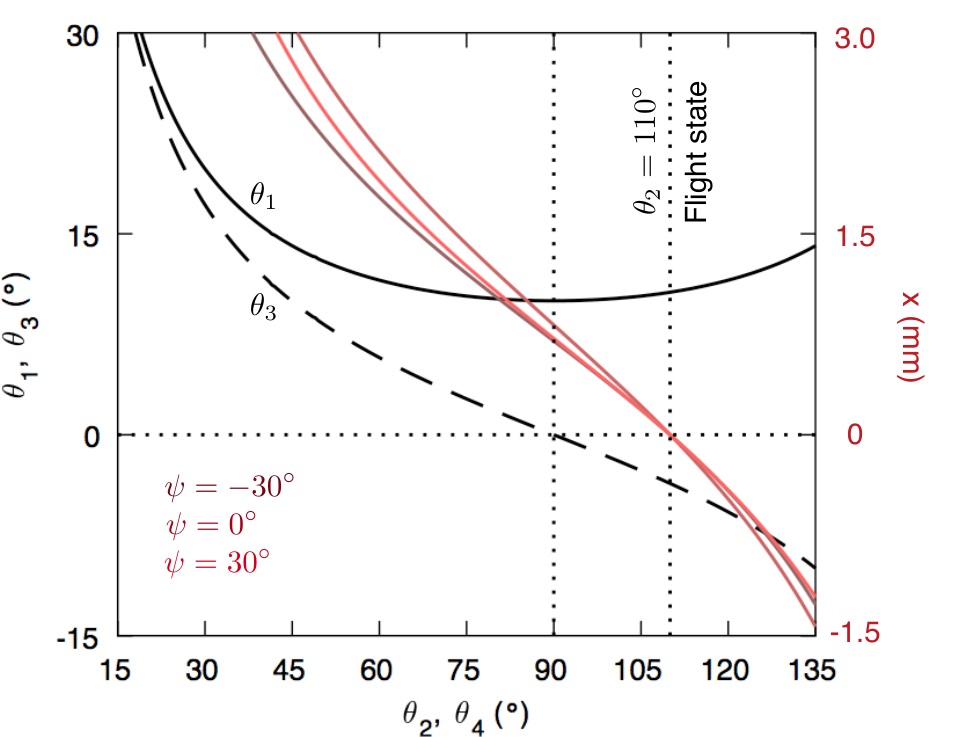}
\caption{Joint kinematics of the 1-DOF foldable airframe. Black lines represent the joint angles at various configurations. The red lines (associated with the red axis on the right) indicates the displacement of the fold trigger.}
\label{fig:kinematics}\vspace{-2mm}
\end{figure}

Each arm is composed of four rigid fiberglass tiles linked together by four flexural hinges as illustrated in Fig.  \ref{fig:panel_folding_arm}B. Tile \textcircled{\footnotesize{1}} is directly connected to the robot's base and acts as a mechanical ground, whereas a motor is attached to the tip of tile \textcircled{\footnotesize{2}}. To achieve the desired kinematics, the exposed Kapton is folded radially (pleat fold) and adhered to the fiberglass. This permanently transforms the planar laminate into a three-dimensional component. \textcolor{black}{This pleat fold creates a two-stage arm folding mechanism (Fig. \ref{fig:panel_folding_stages}) required for the self-locking function in the flight configuration when thrust is generated.} The resultant arm forms a closed-link structure with four revolute joints, described by the angles $\theta_i$'s as indicated in Fig. \ref{fig:panel_folding_arm}B (the rotation directions follow the right hand rule). The motion is constrained to one degree of freedom (DOF). Thanks to the symmetry, $\theta_2$ is always equal to $\theta_4$. The corresponding joint kinematics can be computed using the homogeneous transformation matrices. We define each angle to be zero when its two associated tiles are co-planar. Fig. \ref{fig:kinematics}B shows how $\theta_1$ and $\theta_3$ are related to  $\theta_2$ and $\theta_4$.

\subsection{Folded and Flying States}

The most prominent characteristics of the foldable arm is that the primary fold angle $\theta_1$ has its minimum of $\approx10^\circ$ when $\theta_{2,4}=90^\circ$, or when  tiles \textcircled{\footnotesize{2}} and \textcircled{\footnotesize{3}} are perpendicular. This condition is marked by the \textcolor{black}{left} vertical dotted line in Fig. \ref{fig:kinematics}.

For flight, the motor and propeller are affixed to the far end of tile \textcircled{\footnotesize{2}}. The propeller's thrust produces a positive torque about $\theta_1$. If $\theta_2 > 90^\circ$, this torque undesirably induces further positive rotation of both $\theta_1$ and $\theta_2$. To prevent an overrotation, we implement a joint limit to constrain the rotation of $\theta_2$ to the maximum of $\theta_{2,\text{max}}=110^\circ$ as illustrated in Fig. \ref{fig:panel_folding_arm}C. This designates the equilibrium configuration of the robot in flight. With this design, we passively rely on the propeller's thrust to keep the foldable arm extended when flying. In this configuration, tiles \textcircled{\footnotesize{2}} and \textcircled{\footnotesize{3}}, and, likewise, \textcircled{\footnotesize{1}} and \textcircled{\footnotesize{4}}, are steeply angled. Therefore, they provide the structural rigidity required for flight, impeding the arm from bending in two critical directions.

On the other hand, if we begin with $\theta_2 < 90^\circ$, the positive torque about $\theta_1$ from the thrust results in a negative torque on $\theta_2$. This unfolds $\theta_2$, or the joint between tiles \textcircled{\footnotesize{2}} and \textcircled{\footnotesize{3}}, and folds up $\theta_1$, rendering the arm to be in the folded state. Exploiting another surface of the joint limit located on tile \textcircled{\footnotesize{1}}, the upper limit of $\theta_1$ is physically confined to $70^\circ$. In this condition, the folded arm protects the flight controller or other components on the base from impact. The transition between the flight state and the folded state is achieved using the fold trigger as described later in Section \ref{sec:fold_trigger}.

\subsection{Fold Coupler}
\begin{figure}[t]
\centering
\includegraphics[width=6.0cm]{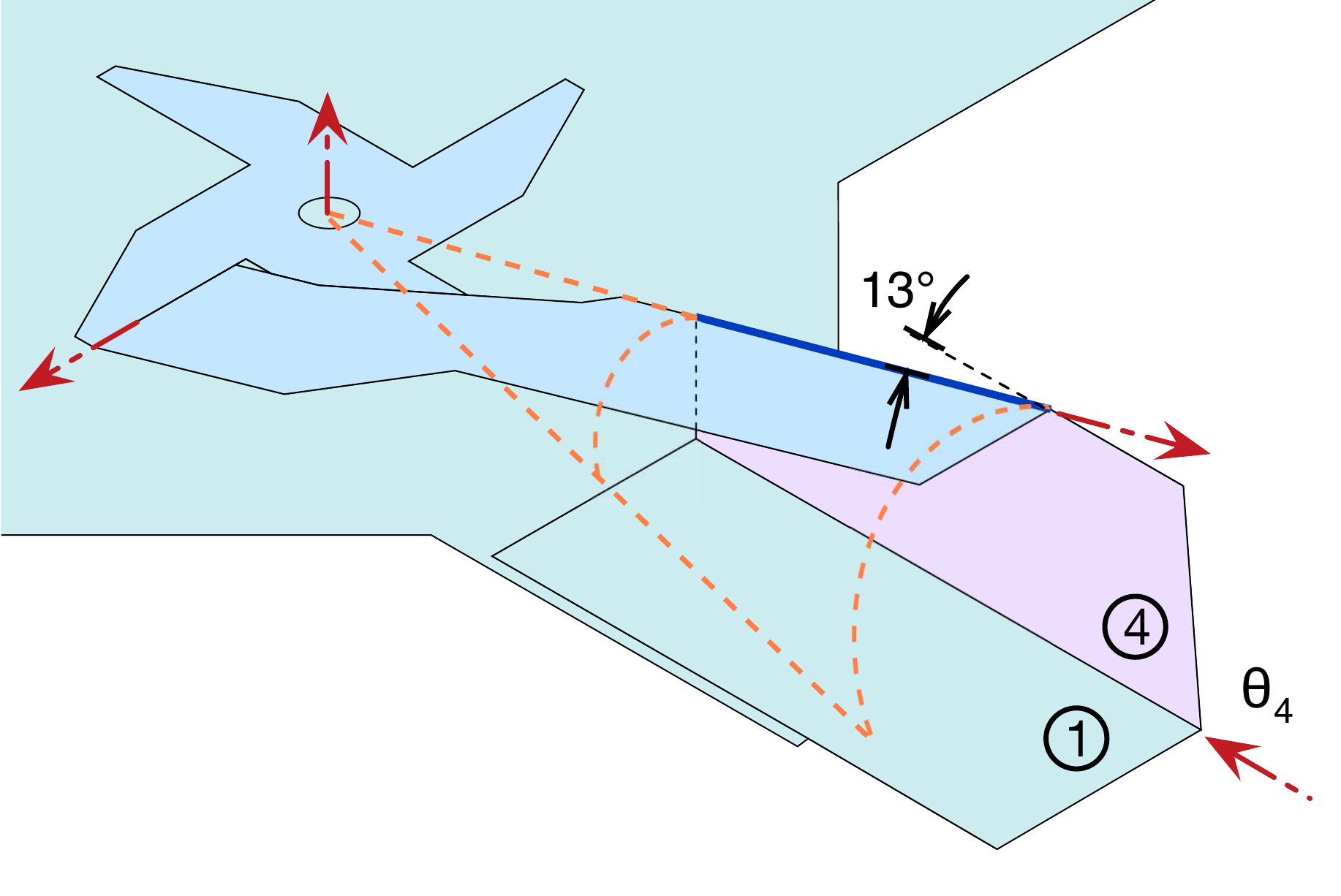}
\caption{A schematic diagram demonstrating the kinematics of the fold coupler and the arm. For clarity, other arms are not shown.}
\label{fig:fold_coupler}
\end{figure}
To ensure all arms fold in synchronization when triggered, we incorporate a fold coupler into the airframe. The coupler is connected to tiles \textcircled{\footnotesize{4}} of all arms. In an ideal condition, this reduces the DOF of the whole airframe to one and the fold angles of all arms are always identical.

The coupling mechanism is implemented into the design of tile \textcircled{\footnotesize{4}}. As depicted in Fig. \ref{fig:fold_coupler}, the inner edge of tile \textcircled{\footnotesize{4}} is tapered by having the top edge trimmed by $13^\circ$. In the folding motion, the trajectory of the top edge (highlighted in navy) forms a surface of an imaginary cone (dashed orange lines). The cone axis coincides with the joint axis of $\theta_4$. The apex of the cone is situated at the center of the airframe. From the top view, the top edge is always radially aligned towards the airframe's center, with the exact position depending on $\theta_4$.

The coupler design leverages the resultant radial symmetry. The coupler is a symmetric tile with a vertical pin joint located at the center of the airframe. The coupler restricts the projected (top view) angles between the top edges of tiles \textcircled{\footnotesize{4}} from multiple arms to $90^\circ$. In the arm folding, all the top edges, and the coupler rotates together as seen from the top. Two additional flexural joints are implemented for each arm to satisfy the associated kinematic constraints in the three dimensional space.

\subsection{Fold Trigger} \label{sec:fold_trigger}
\begin{figure}[t]
\centering
\includegraphics[width=8cm]{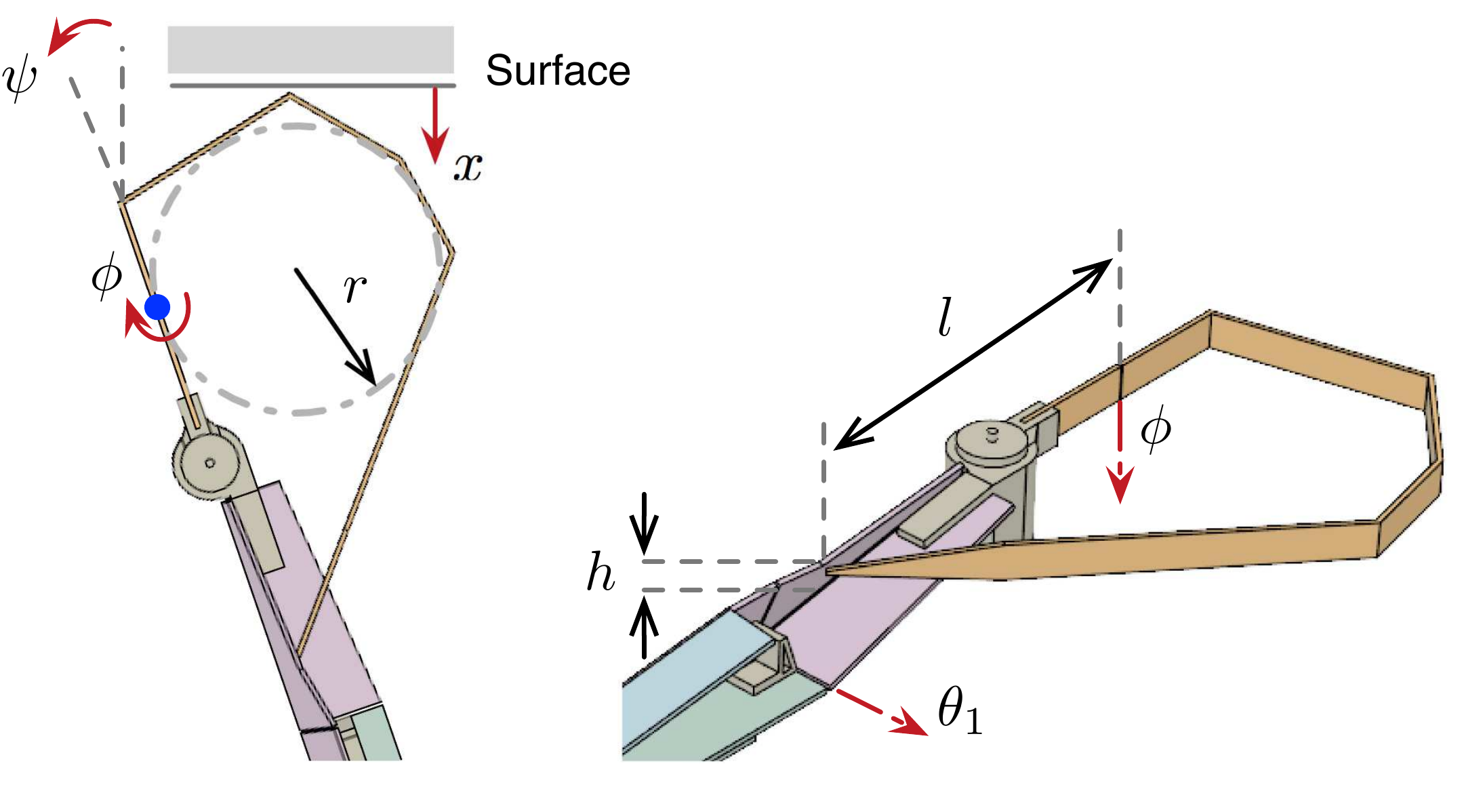}
 \caption{(Left) A top view of the fold trigger and the robot's arm when they collide with a vertical surface (a propeller not shown). The definitions of the yaw angle ($\psi$), the trigger joint ($\phi$), and the displacement ($x$) are given. (Right) An isometric view of the arm and the trigger.}
\label{fig:panel_fold_trigger}
\end{figure}
The transition from the flight state to the folded state upon a collision is obtained via a fold trigger. Fig. \ref{fig:panel_fold_trigger} presents the trigger as a mechanical extension of the arm from the motor mount and tile \textcircled{\footnotesize{2}}. The trigger, also fabricated by lamination, contains one flexural joint ($\phi$) and a tip that makes a contact against tile \textcircled{\footnotesize{3}} when $\phi>0^\circ$. In the nominal flight condition ($\theta_2=\theta_{2,\text{max}}=110^\circ$), $\phi=0^\circ$.

To describe a collision and the folding process, we define the yaw angle ($\psi$) of the robot to represent the heading, or the relative orientation between the robot and surface as shown in Fig. \ref{fig:panel_fold_trigger}. Neglecting small pitch and roll rotations, $\psi=0$ corresponds to the scenario where the arm is perpendicular to the surface. 

Upon impact, the surface causes a positive $\phi$ rotation. The linear displacement of the trigger in the direction perpendicular to the surface ($x$) is directly related to  $\phi$. To simplify the analysis, we regard part of the trigger as a circular arc of radius $r=18$ mm as shown in Fig. \ref{fig:panel_fold_trigger}. As a result, it can be shown that
\begin{equation}{\label{eq:fold_trigger_displacement}}
x = r \left ( \sin(\psi)-\sin(\psi-\phi) \right ).
\end{equation}
The rotation $\phi$ pushes the tip of the trigger against tile \textcircled{\footnotesize{3}} at the distance $h=5$ mm above the joint axis of $\theta_2$ \textcolor{black}{as presented in Fig. \ref{fig:panel_fold_trigger} (Right)}. This loosely couples $\phi$ and $\theta_2$ such that, when the trigger's tip and tile \textcircled{\footnotesize{3}} are in contact ($x>0$, $\phi>0^\circ$), their kinematics satisfy
\begin{equation}{\label{eq:fold_trigger_arm_angle}}
l\sin(\phi) = h \left ( \cot(\theta_2)-\cot(\theta_{2,\text{max}}) \right ),
\end{equation}
where $l=40$ mm is the distance between the contact point to the joint axis of $\phi$ as shown in Fig. \ref{fig:panel_fold_trigger}. By combining equations \eqref{eq:fold_trigger_displacement} and \eqref{eq:fold_trigger_arm_angle}, we can get rid of $\phi$ and numerically obtain a direct relationship between $\theta_2$ and $x$. The result depends on the yaw angle $\psi$. The outcomes are presented in Fig. \ref{fig:kinematics} for three representative yaw angles: $\psi=-30^\circ,0^\circ,30^\circ$. Owing to the symmetry of the robot, the range of $90^\circ$, i.e., $\psi\in(-45^\circ,45^\circ]$, covers all the possibilities.
\begin{figure}[t]
\centering
\includegraphics[width=7.0cm]{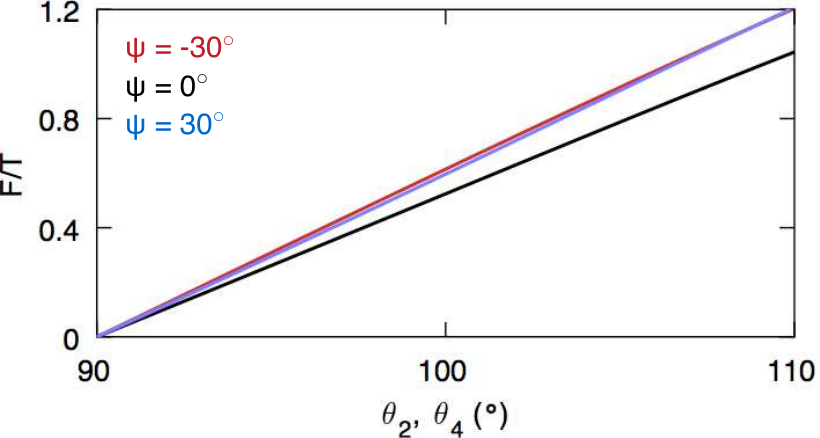}
\caption{The ratio of balanced impact force to thrust ($F/T$) at different arm configurations ($\theta_2$, $\theta_4$) and yaw angels according to equation \ref{eq:trigger_balanced_torque}.}
\label{fig:force_ratio_plot}
\end{figure}

With all the kinematics determined, we proceed to evaluate the collision force required to activate the fold by equating the virtual work done by the propeller's thrust against the force exerted by the surface on the trigger arm ($F$). Since all arms are coupled, the total thrust contributed by all propellers ($T$) must be considered. This thrust acts at the distance $d$ away from the axis of joint $\theta_1$ (with an offset angle $\gamma=10^\circ$, see Fig. \ref{fig:panel_folding_arm}C). This yields $F\mathrm{d} x = -Td\cos\gamma\mathrm{d} \theta_1$ or
\begin{equation}{\label{eq:trigger_balanced_torque}}
\frac{F}{T} = -d\cos\gamma \frac{\mathrm{d}\theta_1}{\mathrm{d}x},
\end{equation}
where $d=40$ mm by design. From the joint kinematics used to produce Fig. \ref{fig:kinematics} and the relationship between $\theta_2$, $x$, and $\psi$ obtained earlier, we numerically compute the force ratio ($F/T$) at various fold configurations and yaw angles. The results, representing the force required for activating the fold at different operating states, are shown in Fig. \ref{fig:force_ratio_plot}. This reveals that the force ratio is an increasing function of $\theta_2$, irrespective of the yaw angle. The activation force $F$ required to initiate the fold when the robot is in the flight state ($\theta_2=110^\circ$) can be found at the upper limit of Fig \ref{fig:force_ratio_plot} when $\theta_2=\theta_{2,\text{max}}$. As previously mentioned, without the pushing force from the trigger, the arm retains its torque equilibrium thanks to the counter torque provided by the joint limit. 

In flight, upon the collision to the wall, the condition for the robot to fold depends on the total thrust. The magnitude of the corresponding impulsive force is determined by equation \eqref{eq:trigger_balanced_torque}. In practice, the force is subject to the impact velocity, the surface properties (such as the coefficient of restitution), etc. Furthermore, we may calculate the amount of work required for activating the fold by integrating equation \eqref{eq:trigger_balanced_torque} over $\mathrm{d}x$, starting from the hovering state ($\theta_2=110^\circ$) to the flipping point ($\theta_2=90^\circ$):
\begin{equation}{\label{eq:trigger_energy}}
W = \int_{\theta_2=110^\circ}^{\theta_2=90^\circ}F\mathrm{d}x = \int_{\theta_2=110^\circ}^{\theta_2=90^\circ} -Td\cos\gamma \frac{\mathrm{d}}{\mathrm{d}x}\theta_1\mathrm{d}x.
\end{equation}
This can be considered the minimum amount of kinetic energy needed in the collision for the robot to fold upon an inelastic collision. This number is independent of the yaw direction. In practice, however, with frictional or viscous losses, and structural compliance, we anticipate the robot would require more energy to overcome the folding process. \textcolor{black}{Other conditions, such as, the rotation of the robot due to the torque induced by the collision, a slippage at the point of impact, etc., may also contribute to inaccurate model predictions.}
\begin{figure}[t]
\centering
\includegraphics[width=8.0cm]{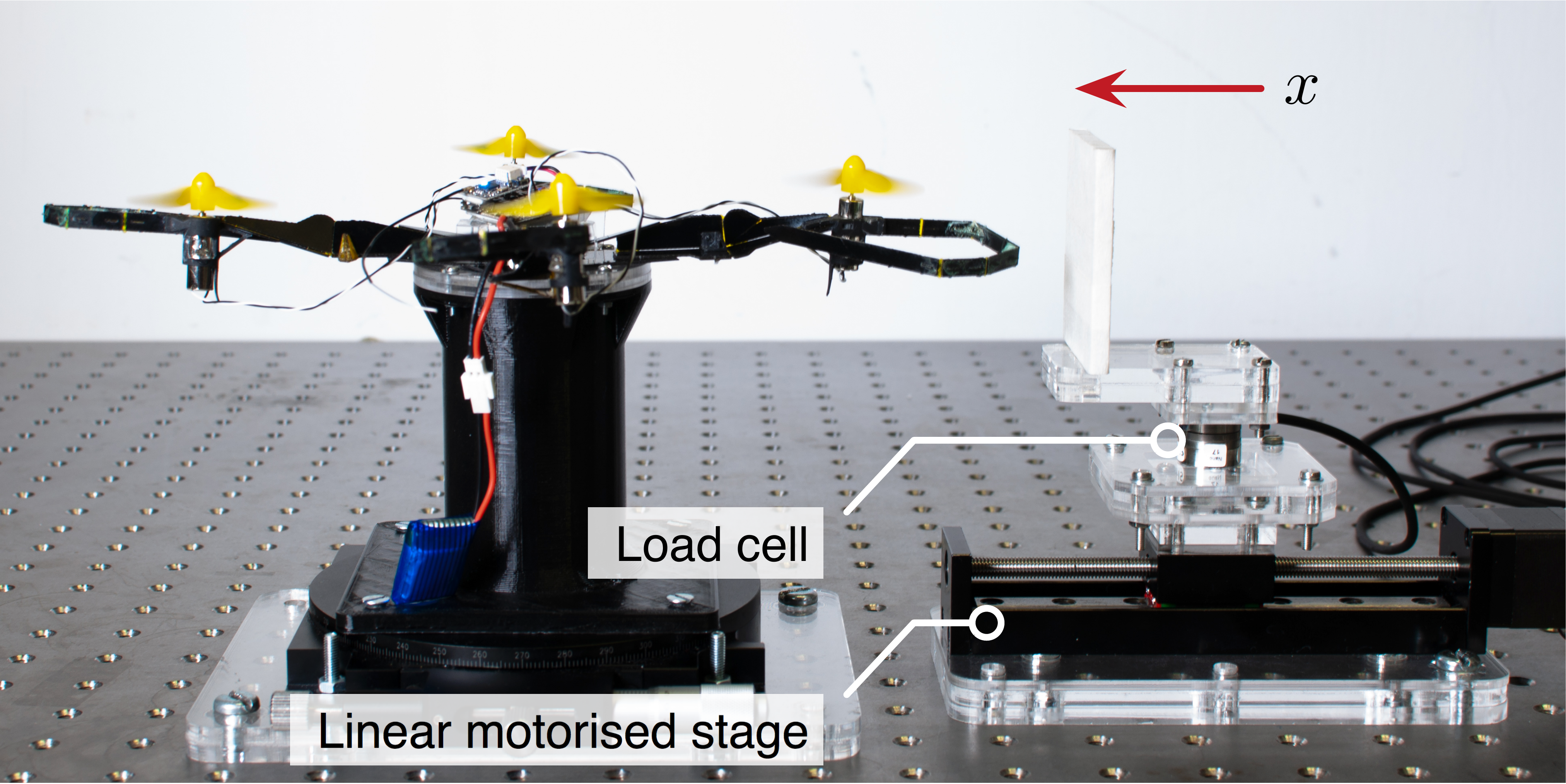}
\caption{Photo of the benchtop experimental setup showing the robot next to the artificial wall. The wall is fixed on a loadcell and a linear stage for force and distance measurements.}
\label{fig:benchtop_setup}
\end{figure}
\section{Benchtop Force Measurements} \label{sec:benchtop_experiments}
In this section, we tested the fabricated robot on a benchtop platform to verify that i) all the arms fold as intended when one of the fold trigger is pushed; and ii) the pushing force and energy required for folding at different yaw angles follows the trend predicted by the models from Section \ref{sec:design_and_model}.
\subsection{Experimental Setup}
\begin{figure*}[tbph]
\centering
\includegraphics[width=17.0cm]{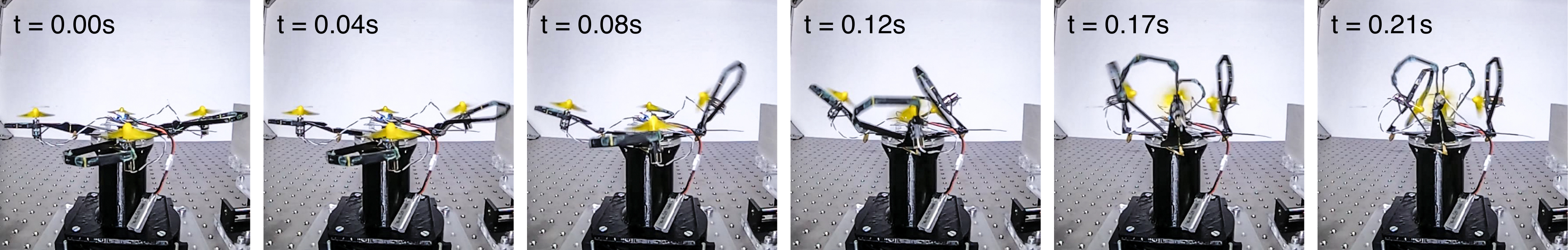}
\caption{Sequential images from a video footage showing the folding process of all four arms when the right arm was pushed by the translating wall. Afterwards, the rigid arms enclosed the central part of the robot for protection from further damage.}
\label{fig:panel_benchtop_fold}\vspace{-2mm}
\end{figure*}
\begin{figure}[t]
\centering
\includegraphics[width=7.0cm]{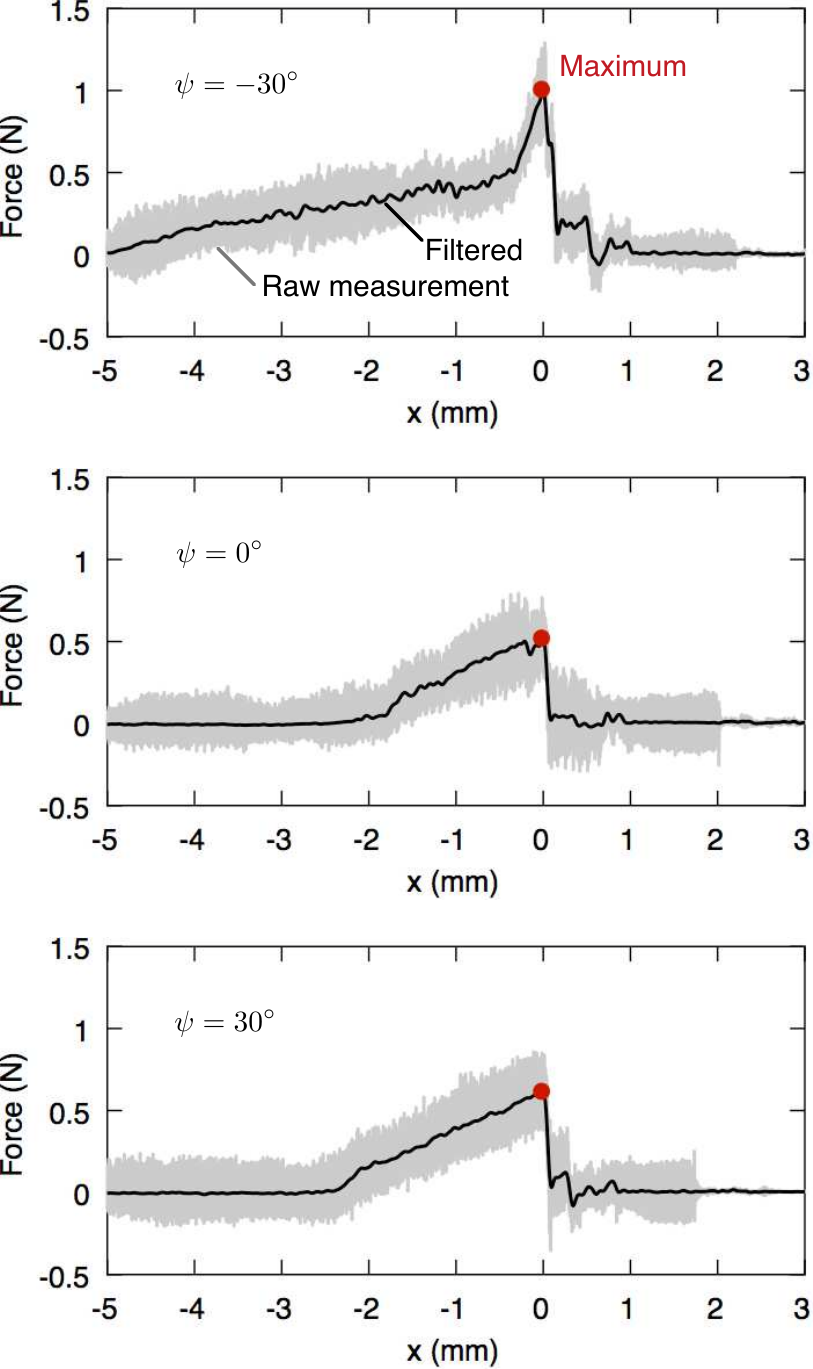}
\caption{Example force measurements plotted against the location of the surface. Three representative data points (out of 68) from three yaw angles are shown. The dark solid lines are filtered measurements and the red dots are the maximum values.}
\label{fig:panel_raw_measurements}\vspace{-2mm}
\end{figure}
\begin{figure}[t]
\centering
\includegraphics[width=6.4cm]{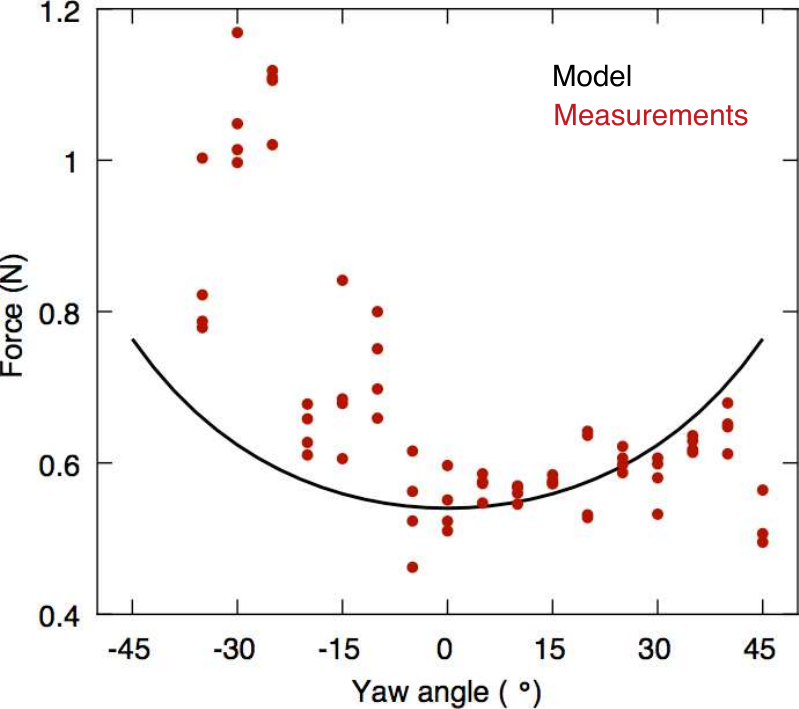}
\caption{The maximum values of measured force from all 68 data points taken at different yaw angles (red dots) in the benchtop experiments, overlaid by the model prediction (black line).}
\label{fig:force_vs_yaw}
\end{figure}
\begin{figure}[t]
\centering
\includegraphics[width=6.4cm]{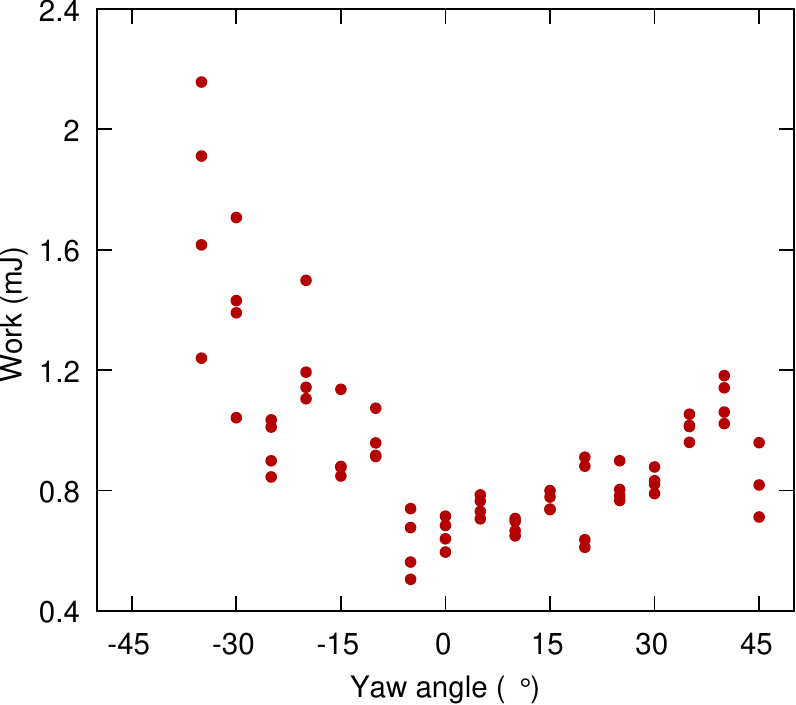}
\caption{The empirical values of energy needed (work done) to fold the airframe in the benchtop experiments.}
\label{fig:work_vs_yaw}\vspace{-2mm}
\end{figure}
To measure the force, the robot is mounted on a 1-DOF rotational platform, enabling a quick and precise adjustment of the yaw angle as illustrated in Fig. \ref{fig:benchtop_setup}. We constructed a vertical rigid surface from an acrylic plate to simulate the collision to a wall. The acrylic plate is placed on top of a load cell (nano17, ATI) and mounted on a linear motorized stage (range 100 mm). The stage was driven by a microstepping driver (TB6600) to translate the surface towards the robot at 0.25 mm.s$^{-1}$ instead of moving the robot towards a static surface. As the force measurements were taken from the surface, not the robot, this reduces measurement noises caused by the vibration from the spinning propellers.

The signal generation for the microstepping driver and the data acquisition were carried out using a computer running the Simulink Real-Time system (Mathworks) with a DAQ (PCI-6229, National Instruments). The experiments were performed when the motors were supplied with 3.7V power. We used the same load cell on a similar setup and equipment to measure the total thrust generated by the robot with the same power supply in advance. The total thrust was found to be $T=0.52$ N, similar to the weight of the robot (0.50 N), or the expected thrust in stable flight.

We performed the force measurements at various yaw angles, ranging from $-35^\circ$ to $45^\circ$ at the increment of $5^\circ$. At each angle, four measurements were taken, making up 68 measurements in total.  The missing angles ($-45^\circ$ and $-40^\circ$) are due to the actual geometry of the fold trigger. In this range, two arms are likely to contact the surface together (at slightly different time), complicating the measurement process. \textcolor{black}{The airframe and flexural joints (Kapton) did not break or tear over the course of the experiments.}

\begin{figure*}[th]
\centering
\includegraphics[width=17.0cm]{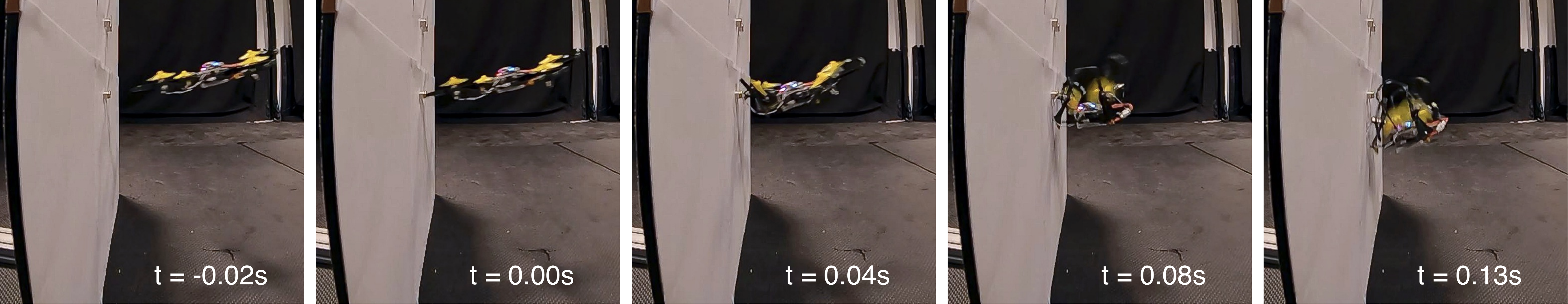}
\caption{Sequential images from a video footage showing the mid-flight collision when the robot had the translational speed of $1.5$ m.s$^-1$. The images reveal that the impact triggered the folding of all arms in less than 0.15 s.}
\label{fig:panel_flight}\vspace{-2mm}
\end{figure*}

\subsection{Measurement Results}

Fig. \ref{fig:panel_benchtop_fold} shows an example of the folding process when the fold trigger was pushed by the moving surface as recorded by a camera at 240 Hz. The image sequence reveals all the arms started to fold at slightly different time and the whole fold completes in approximately $0.2$ s. The observed asynchronization is expected in practice due to the inherent structural compliance of parts and flexural hinges that were neglected in the analysis of the kinematics. Nevertheless, this verifies that the coupler functions as intended. \textcolor{black}{After folding, the propellers were more than one centimeter from the control board. The gap size was deemed sufficiently large as a result of the carefully chosen design parameters. It is possible to further reduce the risk of propeller-controller contact by incorporating magnetic joints to automatically power down the propellers as in \cite{mintchev2017insect}, with a trade-off of the added weight and design complexity.}

Three representative force measurements with respect to the wall position are plotted in Fig. \ref{fig:panel_raw_measurements}. Raw force measurements are low-pass filtered (cut off frequency of 50 Hz) and the maximum filtered forces are marked by the red dots. The corresponding positions are labelled $x=0$ mm. It can be seen that, prior to this point, the force increases with $x$.

As the translating platform moves towards the robot, but no contact is made ($x$ is negative), the measured force is approximately zero. Upon contact, the surface pushes the fold trigger, producing the torque countering the torque from the thrust, resulting in non-zero force measurements. This incrementally and simultaneously replaces the counter torque contributed by the joint limit. The process occurs over a non-zero distance due to the unmodeled compliance of the airframe. At the maximum, the pushing force completely overcomes the thrust. It is reasonable to assume that the ratio of the maximum $F$ to $T$ corresponds to force ratio ($F/T$) when $\theta_2=\theta_{2,\text{max}}=110^\circ$ presented in Fig. \ref{fig:force_ratio_plot}.

The maximum measured forces from all experiments are plotted against the yaw angle in Fig. \ref{fig:force_vs_yaw}. Also shown is the prediction from the model as given by equation \eqref{eq:trigger_balanced_torque}, with $T=0.52$ N. The results show a reasonable agreement with the model for positive yaw angles. However, the measurements are up to $\approx30\%$ higher than the model prediction for some negative yaw angles. We believe the discrepancy is caused by several factors. One explanation is the simplification in the model that treats the geometry of the fold trigger as a circular arc. This could lead to an incorrect point of contact to the wall, contact angle, and different effective moment arms, all of which possibly contribute to the modeling errors. Another aspect is the inherent compliance of the structure and flexural joints, together with friction, structural vibration, and damping effects that are not considered in the model.

In addition to the force required to activate the fold, we compute the amount of energy or work exerted by the surface by numerically integrating the force over displacement (corresponding to the areas covered in Fig. \ref{fig:panel_raw_measurements}) for all data points. The results are given in figure \ref{fig:work_vs_yaw}. The plot reveals a similar trend to the measured maximum force in Fig. \ref{fig:force_vs_yaw}. At most angles, the robot required $\approx 0.5-1$ mJ to fold, with the exception near $\psi\approx -45^\circ$. The obtained values are generally a few times larger than the theoretical prediction given by equation \eqref{eq:trigger_energy} of $0.23$ mJ (when $T=0.52$ N). This is not surprising as the theoretical bound does not take into account frictional losses and compliance in the structure and flexural joints. It is likely that these unaccounted effects are more pronounced at negative yaw angles, resulting in large values of force and work as observed in Fig. \ref{fig:force_vs_yaw} and \ref{fig:work_vs_yaw}.

In a flight scenario, if it is assumed that the kinetic energy is all taken up as work required for the robot to fold upon the collision to a wall, the amount 2 mJ (taken from Fig. \ref{fig:work_vs_yaw}) equates to the impact speed of $\approx 30$ cm.s$^{-1}$ for the robot with 51.2g mass. It is conceivable that with other losses or if not all energy is converted, the minimum speed required for activating the fold could be higher.

Overall, the experiments verify that the proposed mechanism enables all four arms to fold when one arm is triggered. This is achieved over almost 360$^\circ$ of yaw angle in a static scenario (taking into account the symmetry of the robot). Moreover, the measurement results suggest that, depending on the yaw angle, the force required is in the same order of magnitude as the weight of the robot. In other words, no significant impact is needed for the fold activation.

\section{Mid-Flight Collision Demonstration} \label{sec:mid-flight_collision}
With the battery, onboard electronics, and  the commercial flight controler, the quadrotor is capable of stable flight. To demonstrate the fold activation from a mid-flight collision, the robot was remotely controlled to fly horizontally at the speed of $\approx$1.5 m.s$^{-1}$ (estimated from the video footage) towards a vertical surface (acrylic plate covered by paper for improved visibility). This speed is notably above the calculated bound of 30 cm.s$^{-1}$. Upon impact, the airframe completed the fold in less than $0.2$ s before crashing to the ground in the folded configuration. The flight was captured by a camera at 240 Hz. The video frames are shown in Fig. \ref{fig:panel_flight}. The foldable structure and onboard components were intact after the crash.

\section{Conclusion \textcolor{black}{and Discussion}}
This paper employs an origami-inspired strategy for a small multicopter to protect the delicate components residing in the center of the body from crashing to the ground after a collision. This is achieved by creating a foldable structure that is activated by impact, allowing the airframe to deform in a mid-flight collision such that the rigid structure shields the body part from the fall. The proposed design has been experimentally verified in both static measurements and actual flights.

\textcolor{black}{Despite the successful demonstration in flight, the proposed design and modeling efforts still have several restrictions. In addition to the assumptions on the collision conditions employed in section \ref{sec:fold_trigger}, the proposed mechanism does not protect the robot from top or bottom collisions. While a full protection is present in a caged design \cite{kornatowski2017origami}, the limited protection is also a shortcoming of the protective mechanisms in \cite{mintchev2017insect,sareh2018rotorigami}}.

In summary, the developed prototype makes use of an intelligent mechanical design to overcome the contradicting requirements on the structural stiffness. By aligning a pair of planar structures in a perpendicular direction, we obtain the desired rigidity. The interlocking mechanism, or joint limits, lets the robot use the thrust force to remain in the flight state. The fold is passively triggered as the impact force overcomes the thrust. The activation, as a result, does not necessitate an extra sensor or actuator. We believe the presented solution can be further extended for other applications, in particular, for small flying robots that severely suffer from restrictions in payload and power consumption.

\subsection{\textcolor{black}{Origami-Inspired Design and Scalability}}

\textcolor{black}{As shown in the mid-flight collision, the proposed origami-inspired airframe  not only allows the robot to reconfigure when triggered, but also enables the use of thin planar materials to provide sufficient rigidity required for flight. A single sheet of 300-$\mu$m fiberglass alone would bend significantly when subject to the propeller's thrust ($\approx0.13$ N), but when two sheets are oriented perpendicularly, the effective stiffness increases dramatically. This strategy has been refered to as an origami principle of perpendicular folding in \cite{kim2018origami}.}

\textcolor{black}{Fundamentally, our foldable mechanism is constructed from two materials with vastly different stiffness (fiberglass for tiles and Kapton for flexural joints).  The condition is approximately equivalent to rigid panels connected by frictionless hinges. According to \cite{schenk2011origami}, the behavior of the foldable structure is predominantly determined by its geometry as opposed to the exact material properties. This is valid irrespective of size, rendering the origami-inspired design effective at various scales \cite{rus2018design}, from micrometers \cite{whitney2011pop} to meters \cite{kim2018origami}. For these reasons, to the first order, the bending of fiberglass panels can be neglected in the model. In addition to this, the discrepancy between the model predictions and experimental results in this work is likely contributed by the frictional losses and vibrations.}

\textcolor{black}{In regard to the scalability of the required triggering force, equation \eqref{eq:trigger_balanced_torque} suggests that the ratio of the trigger force ($F$) to the robot's weight ($T$) is independent of size as both $d$ and $x$ on the right hand side of equation \eqref{eq:trigger_balanced_torque} are expected to scale with the characteristic length ($l$) of the robot. In other words, we anticipate $F\sim T$ or $F\sim l^3$. It follows that the required impact energy upon a collision, given by equation
\eqref{eq:trigger_energy} scales as $W\sim\int F\mathrm{d}x\sim l^4$. In the meantime, the kinetic energy of the robot ($K$) is $K=\frac{1}{2}mv^2$, when $v$ is the speed at impact. Assuming $v\sim \sqrt{l}$ as found for flying machines and other modes of locomotion \cite{bejan2006unifying}, then $K\sim l^4$. With the assumption that the energy needed to trigger the fold is proportional to the kinetic energy at impact, the scaling analysis suggests that the fold trigger mechanism is applicable for different vehicle sizes as $W\sim K\sim l^4$. Nevertheless, as robots are scaled up and the payload and flight endurance become less restricted \cite{karydis2017energetics},  a foldable mechanism becomes a less compelling approach as conventional solutions involving sensors and actuators or a protective cage can be employed instead.}

\bibliography{bibtex}
\bibliographystyle{IEEEtran}


\end{document}